\documentclass[runningheads]{llncs}

\usepackage[american]{babel}
\usepackage{notations}
\usepackage{microtype}
\usepackage{graphicx, xcolor}
\usepackage{afterpage}
\usepackage{subfigure}
\usepackage{amsmath,amsfonts,amssymb,bm,bbm,mathptmx,dsfont}
\usepackage[mathscr]{euscript}
\usepackage{booktabs} 
\usepackage{url}
\usepackage{array}
\usepackage[font=small]{caption}
\usepackage{arydshln}
\usepackage{booktabs} 
\usepackage{mathtools} 
\usepackage[round]{natbib}

\definecolor{Blue}{rgb}{0.1, 0.1, 0.5}
\usepackage{hyperref}
\hypersetup{bookmarksnumbered,
            colorlinks=true,
            citecolor=Blue,
            linkcolor=black,
            breaklinks,
            pdfborder=0 0 0,
            pdftitle=Numerical issues in maximum likelihood parameter estimation for Gaussian process regression,
            pdfauthor=E. Vazquez,
            pdfkeywords=Gaussian process regression; maximum likelihood}

\newcommand{\mysubsec}[1]{%
  \vspace{-0.3em}\subsection{#1}\vspace{-0.05em}}
\newcommand{\mysec}[1]{%
  \vspace{-0.5em}\section{#1}\vspace{-0.5em}}
\newcommand{\mypar}[1]{%
  \medbreak\noindent\emph{#1}\ }
\newcommand{\EOFvspace}{\vskip -5mm}

\begin{document}
\title{Numerical issues in maximum likelihood parameter estimation for Gaussian process interpolation}
\titlerunning{Maximum likelihood issues in Gaussian process interpolation}
\author{
Subhasish Basak\inst{1} \and
Sébastien Petit\inst{1,2} \and
Julien Bect\inst{1}\and
Emmanuel Vazquez\inst{1}}
\authorrunning{Subhasish Basak et al.}
\institute{Laboratoire des Signaux et Systèmes, 
CentraleSupélec, CNRS, Univ. Paris-Saclay, Gif-sur-Yvette, 
France.  \email{<firstname>.<lastname>@centralesupelec.fr} \and
Safran Aircraft Engines, Moissy-Cramayel, France}
\maketitle
\begin{abstract}
This article investigates the origin of numerical issues in
maximum likelihood parameter estimation for Gaussian process (GP)
interpolation and investigates simple but effective strategies for
improving commonly used open-source software implementations.
This work targets a basic problem but a host of studies, particularly
in the literature of Bayesian optimization, rely on off-the-shelf GP
implementations. For the conclusions of these studies to be reliable
and reproducible, robust GP implementations are critical.

\keywords{Gaussian process \and Maximum likelihood estimation \and Optimization.}
\end{abstract}

\section{Introduction}
\label{sec:introduction}

Gaussian process (GP) regression and interpolation \citep[see,
e.g.,][]{rasmussen06:_gauss_proces_machin_learn}, also known as kriging
\citep[see, e.g.,][]{stein1999interpolation}, has gained significant
popularity in statistics and machine learning as a non-parametric
Bayesian approach for the prediction of unknown functions. The need for
function prediction arises not only in supervised learning tasks, but
also for building fast surrogates of time-consuming computations, e.g.,
in the assessment of the performance of a learning algorithm as a
function of tuning parameters or, more generally, in the design and
analysis computer experiments
\citep{santner03:_desig_analy_comput_exper}. The interest for GPs has
also risen considerably due to the development of Bayesian optimization
(\citealp{mockus75:_bayes, Jones1998, emmerich2006, srinivas2010}\ldots).

This context has fostered the development of a fairly large number of
open-source packages to facilitate the use of GPs. Some of the popular
choices are the Python modules scikit-learn \citep{scikit-learn}, GPy
\citep{gpy2012}, GPflow \citep{GPflow2017}, GPyTorch
\citep{gardner2018gpytorch}, OpenTURNS \citep{OpenTURNS}; the R package
DiceKriging \citep{DiceKriging}; and the Matlab/GNU Octave toolboxes GPML
\citep{rasmussen10:_gauss_proces_machin_learn_gpml_toolb}, STK
\citep{STK} and GPstuff
\citep{vanhatalo12:_bayes_model_gauss_proces_matlab}.

In practice, all implementations require the user to specify the mean
and covariance functions of a Gaussian process prior under a
parameterized form. Out of the various methods available to estimate the
model parameters, we can safely say that the most popular approach is
the \emph{maximum likelihood estimation} (MLE) method. 
However, a simple numerical experiment consisting in interpolating a
function (see Table~\ref{table:diff-results}), as is usually done in
Bayesian optimization, shows that different MLE implementations from
different Python packages produce very dispersed numerical results when
the default settings of each implementation are used.  These significant
differences were also noticed by \cite{erickson18:_compar_gauss} but the
causes and possible mitigation were not investigated. Note that each
package uses its own default algorithm for the optimization of the
likelihood: GPyTorch uses ADAM \citep{kingma2015:_adam}, OpenTURNS uses
a truncated Newton method \citep{Nash1984} and the others generally use L-BFGS-B
\citep{byrd1995limited}.  It turns out that none of the default results
in Table~\ref{table:diff-results} are really satisfactory compared to
the result obtained using the recommendations in this
study\footnote{Code available at \url{https://github.com/saferGPMLE}}.

\begin{table}[t]
  \caption{Inconsistencies in the results across different Python
    packages. The results were obtained by fitting a GP model, with
    constant mean and a Matérn  kernel ($\nu=5/2$), to the Branin function,
    using the default settings for each package. We used $50$~training points
    and $500$~test points sampled from a uniform distribution on
    $[-5, 10] \times [0, 15]$.
    The table reports the estimated values for the variance and length
    scale parameters of the kernel, the empirical root mean squared
    prediction error (ERMSPE) and the minimized negative log likelihood
    (NLL). The last row shows the improvement using the recommendations
    in this study.}
  \label{table:diff-results}
\begin{center}
\footnotesize \setlength{\tabcolsep}{6pt}
\begin{tabular}{lc|cc|cr}
\sc{Library} & Version & Variance & Lengthscales & ERMSPE & NLL \\[0.5ex]
\hline%
\rule{0pt}{10pt}%
\sc{scikit-learn } & 0.24.2 & $9.9\cdot 10^4$ & $(13,\, 43)$  & $1.482$ & $132.4$\\
\sc{GPy } & 1.9.9 & $8.1\cdot 10^8$ & $(88,\, 484)$ & $0.259$ & $113.7$\\
\sc{GPytorch } & 1.4.1 & $1.1\cdot 10^1$ & $(4,\, 1)$ & $12.867$ & $200839.7$\\
\sc{GPflow } & 1.5.1 & $5.2\cdot 10^8$ & $(80,\, 433)$ & $0.274$ &  $114.0$\\
\sc{OpenTURNS } & 1.16 & $1.3\cdot 10^4$ & $(8,\, 19)$ & $3.301$ &  $163.1$\\
\hdashline 
\vphantom{$\Bigm|$}%
\sc{GPy ``improved''} & 1.9.9 & $9.4 \cdot 10^{10}$ & $(220,\, 1500)$ & $0.175$ & $112.0$\\[-2pt]
\bottomrule
\end{tabular}
\end{center}
\EOFvspace
\end{table}

\begin{figure}[p]
\begin{center}
\setlength{\subfigbottomskip}{8mm}
\subfigure[optimized NLL]{\includegraphics[trim=20 0 60 15,clip,width=5.8cm]{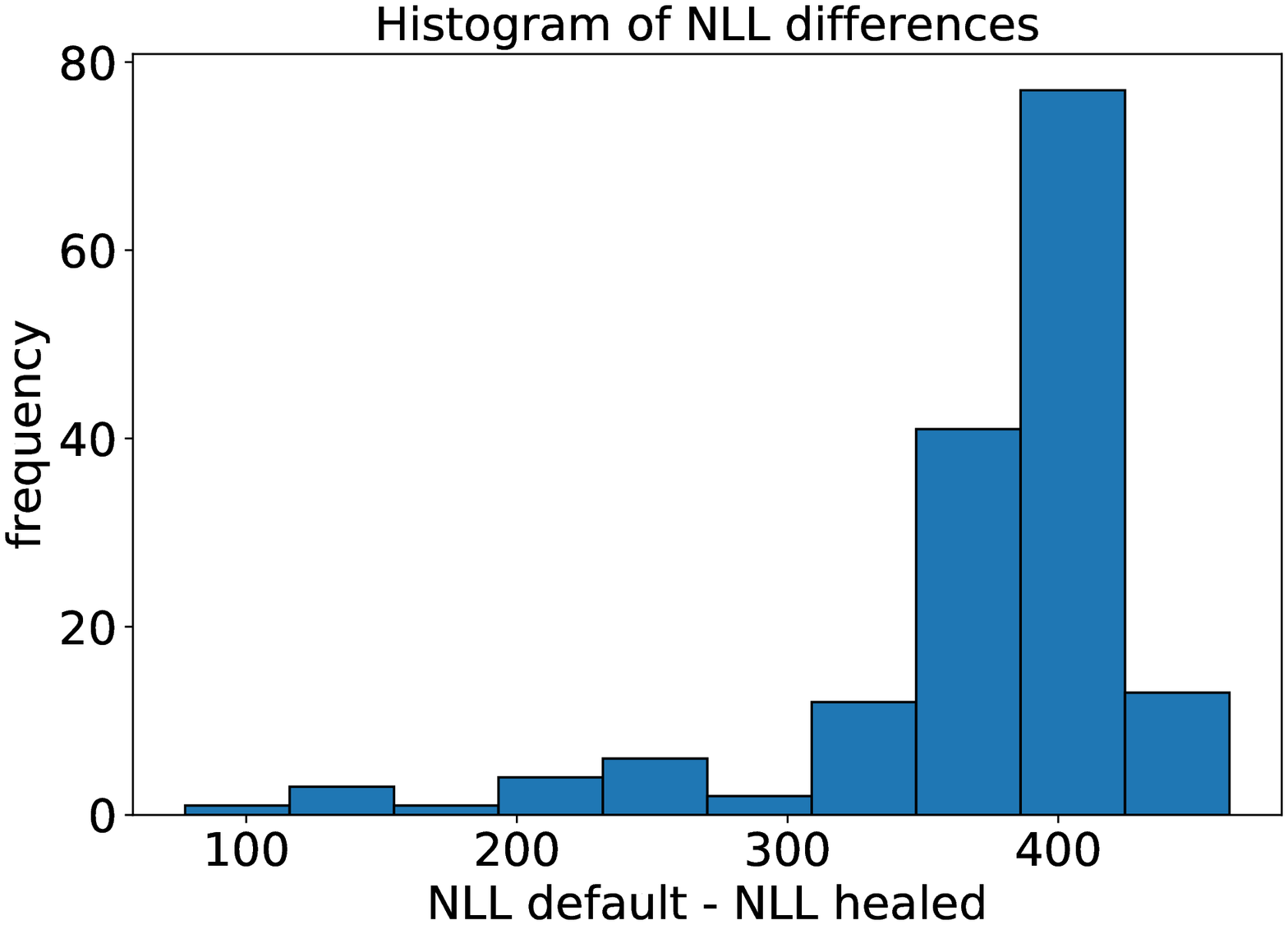}}
\hspace{3mm}
\subfigure[prediction error]{\includegraphics[trim=20 0 60 15,clip,width=5.8cm]{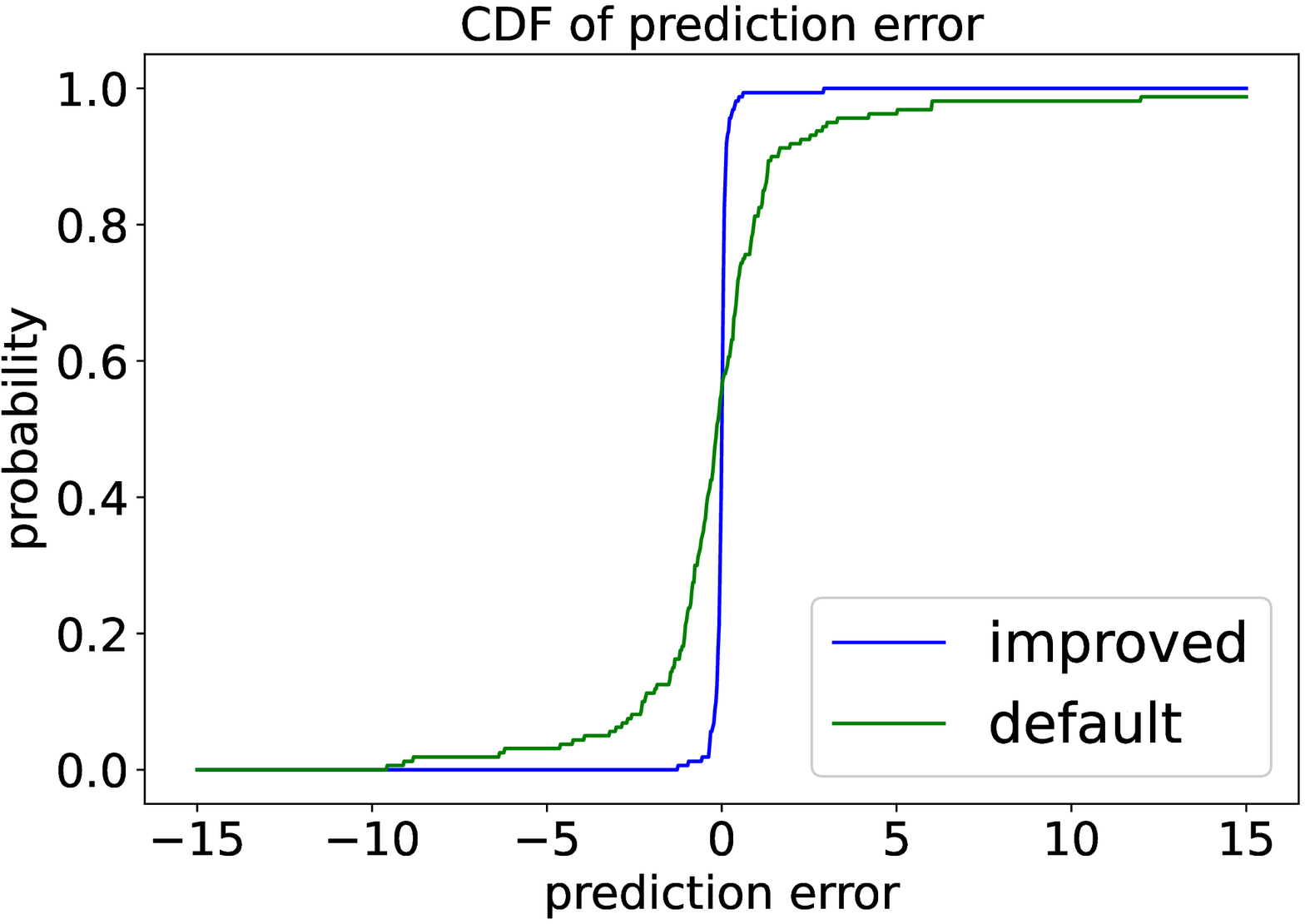}}
\subfigure[optimized lengthscales]{\includegraphics[trim=50 0 90 50,clip,width=1.01\textwidth]{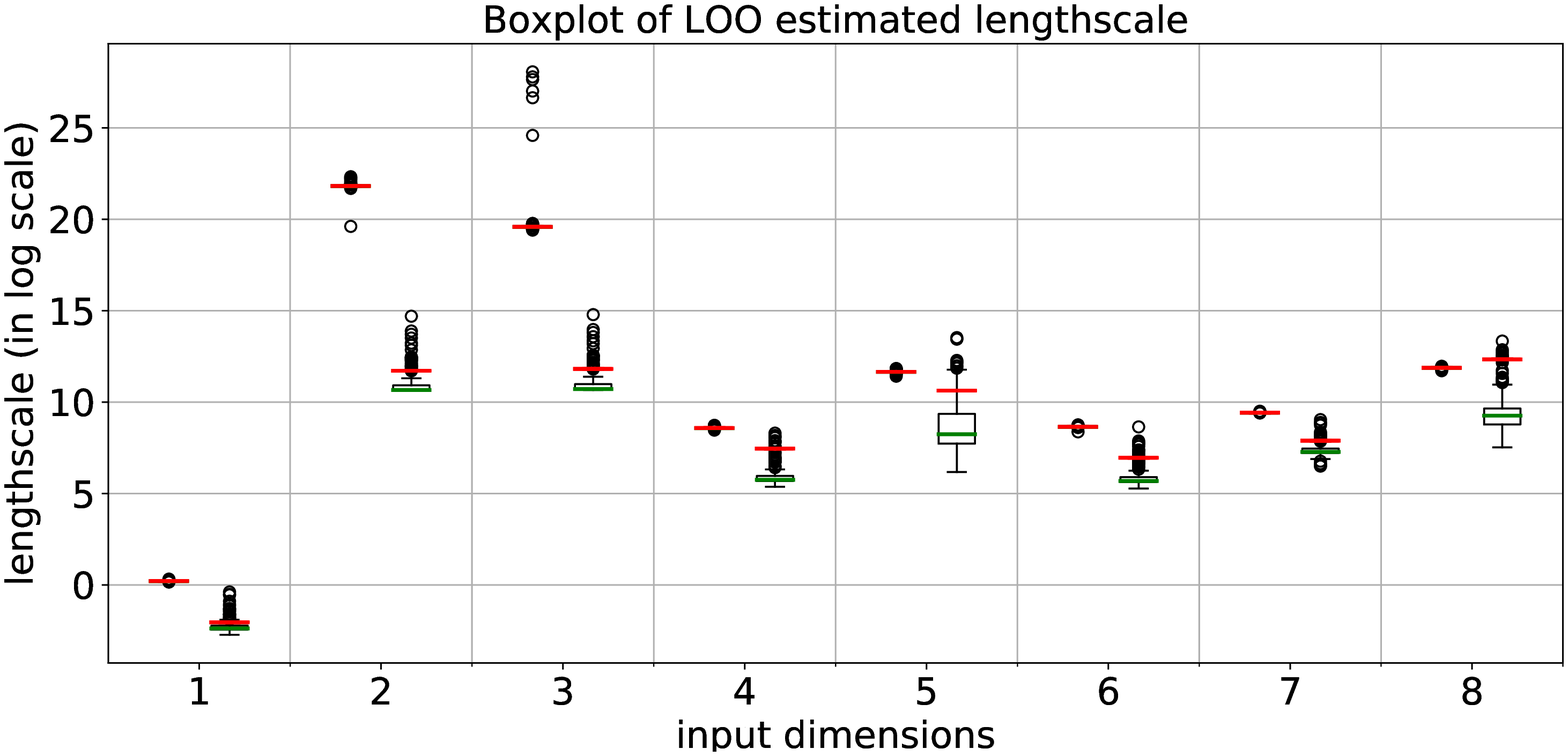}}

\caption{\emph{Improved} (cf.~Section~\ref{sec:conclusions}) vs
  \emph{default} setups in GPy on the Borehole function with
  $n = 20d = 160$ random training points. We remove one point at a time
  to obtain (a) the distribution of the differences of negative
  log-likelihood~(NLL) values between the two setups; (b) the empirical CDFs of
  the prediction error at the removed points; (c) pairs of box-plots for
  the estimated range parameters (for each dimension, indexed from 1 to
  8 on the $x$-axis, the box-plot for
  \emph{improved} setup is on the left and the box-plot for
  \emph{default} setup is on the right; horizontal red lines correspond
  to the estimated values using the whole data set without leave-one-out). Notice that the
  parameter distributions of the \emph{default} setup are more spread
  out.}
\label{fig:LOO-plots}
\end{center}
\end{figure}

Focusing on the case of GP interpolation (with Bayesian optimization as
the main motivation), the first contribution of this article is to
understand the origin of the inconsistencies across available
implementations. The second contribution is to investigate simple but
effective strategies for improving these implementations, using the
well-established GPy package as a case study.  We shall propose
recommendations concerning several optimization settings: initialization
and restart strategies, parameterization of the covariance, etc.  By
anticipation of our numerical results, the reader is invited to refer to
Figure~\ref{fig:LOO-plots} and Table~\ref{table:LOO-mse}, which show
that significant improvement in terms of estimated parameter values and
prediction errors can be obtained over default settings using better
optimization schemes.

Even though this work targets a seemingly prosaic issue, and advocates
somehow simple solutions, we feel that the contribution is nonetheless
of significant value considering the widespread use of GP modeling.
Indeed, a host of studies, particularly in the literature of Bayesian
optimization, rely on off-the-shelf GP implementations: for their
conclusions to be reliable and reproducible, robust
implementations are critical.

The article is organized as follows. Section~\ref{sec:background}
provides a brief review of GP modeling and
MLE. Section~\ref{sec:numerical-aspects} describes some numerical
aspects of the evaluation and optimization of the likelihood function,
with a focus on GPy's implementation. Section~\ref{sec:healing} provides
an analysis of factors influencing the accuracy of numerical MLE
procedures.  Finally, Section~\ref{sec:numerical-experiments} assesses
the effectiveness of our solutions through numerical experiments and
Section~\ref{sec:conclusions} concludes the article.

\begin{table}[t]
  \caption{\emph{Improved} (cf.~Section~\ref{sec:conclusions}) vs
    \emph{default} setups in GPy for the interpolation of the Borehole
    function (input space dimension is $d=8$) with $n\in\{3d,\, 5d\}$
    random data points (see Section~\ref{subsec:datasets} for
    details). The experiment is repeated 50 times. The columns report
    the leave-one-out mean squared error (LOO-MSE) values (empirical mean over the repetitions, together
    with the standard deviation and the average proportion of the
    LOO-MSE to the total standard deviation of the data in parentheses).}
  \label{table:LOO-mse}
\begin{center}
\begin{small}
\begin{tabular}{p{2cm}p{3.4cm}p{3cm}}
\toprule
\sc{Method} 
& $n = 3d$ & $n = 5d$ \\
\midrule
\sc{Default} 
& 17.559~ (4.512, 0.387)
& 10.749~ (2.862, 0.229)
\\ 
\sc{Improved} 
& 3.949~ (1.447, 0.087)
& 1.577~ (0.611, 0.034)
\\
\bottomrule
\end{tabular}
\end{small}
\end{center}
\EOFvspace
\end{table}

\mysec{Background}
\label{sec:background}

\mysubsec{Gaussian processes}
\label{subsec:GP}

Let $Z \sim {\mathrm{GP}}(m,\, k)$ be a Gaussian process
indexed by $\RR^d$, $d\geq 1$, specified by a mean function
$m:\RR^d\to \RR$ and a covariance function $k:\RR^d\times\RR^d\to \RR$.

The objective is to predict $Z(x)$ at a given location $x\in\RR^d$,
given a data set
$D = \{(x_i,\,z_i)\in\RR^d\times\RR,\, 1 \le i \le n\}$, where the
observations $z_i$s are assumed to be the outcome of an additive-noise
model: $Z_i = Z(x_i) + \varepsilon_i$, $1 \le i \le n$. In most
applications, it is assumed that the $\varepsilon_i$s are zero-mean
Gaussian i.i.d.\ random variables with variance
$\sigma_{\varepsilon}^2\geq 0$, independent of $Z$. (In rarer cases,
heteroscedasticity is assumed.)

Knowing $m$ and $k$, recall \citep[see,
e.g.][]{rasmussen06:_gauss_proces_machin_learn} that the posterior
distribution of~$Z$ is such that
$Z \mid  Z_1,\,\ldots,\, Z_n,\, m,\, k \sim GP(\hat Z_n,\, k_n)$,
where $\hat Z_n$ and $k_n$ stand respectively for the posterior mean
and covariance functions:
\begin{equation*}
  \begin{array}{ll}
  \hat Z_n(x) &= m(x) + \sum_{i=1}^n w_i(x;\x_n)\, (z_i - m(x_i))\,,\\[1em]
  k_n(x, y) &=\; k(x,y) - \mb{w}(y;\x_n)\tr \mb{K}(\x_n, x) \,,    
  \end{array}
\end{equation*}
where $\x_n$ denotes observation points
$\left( x_1,\, \ldots,\, x_n\right)$ and the weights $w_i(x;\x_n)$ are
solutions of the linear system:
\begin{equation}
  (\mb{K}(\x_n,\x_n) + \sigma_{\varepsilon}^2 \mb{I}_{n})\, \mb{w}(x;\x_n) = \mb{K}(\x_n, x)\,,
\end{equation}
with $\mb{K}(\x_n,\,\x_n)$ the $n\times n$ covariance matrix with
entries $k(x_i,\,x_j)$, $\mb{I}_n$ the identity matrix of size~$n$, and
$\mb{w}(x;\x_n)$ (resp.~$\mb{K}(\x_n, x)$) the column vector with
entries $w_i(x;\x_n)$ (resp.~$k(x_i, x)$), $1 \le i \le n$.
The posterior covariance at $x,\, y\in\RR^d$ may be written as
\begin{align}
    \label{eq:def-krig-cov}
    k_n(x, y)\, &=\; k(x,y) - \mb{w}(y;\x_n)\tr \mb{K}(\x_n, x) \,.
\end{align}

It is common practice to assume a zero mean function $m = 0$---a
reasonable choice if the user has taken care to center data---but most
GP implementations also provide an option for setting a constant mean function
$m(\,\cdot\,) = \mu\in\RR$. %
In this article, we will include such a constant in our models, and
treat it as an additional parameter to be estimated by MLE along with
the others.  (Alternatively, $\mu$ could be endowed with a Gaussian or
improper-uniform prior, and then integrated out; see, e.g.,
\citet{ohagan78:_curve_fittin_optim_desig_predic}.)

The covariance function, aka covariance kernel, models similarity
between data points and reflects the user's prior belief about the
function to be learned.  Most GP implementations provide a
couple of stationary covariance functions taken from the literature
\citep[e.g.,][]{wendland04:_scatt,rasmussen06:_gauss_proces_machin_learn}. The
\emph{squared exponential}, the \emph{rational quadratic} or the
\emph{Matérn} covariance functions are popular choices (see
Table~\ref{table:kernels}). These covariance functions include a number
of parameters: a variance parameter $\sigma^2>0$ corresponding to the
variance of $Z$, and a set of range (or length scale) parameters
$\rho_1,\,\ldots,\, \rho_d$, such that
\begin{equation}
  \label{eq:cov-gen-form}
  k(x,y) = \sigma^{2}r(h)\,,
\end{equation}
with $h^2 = \sum_{i=1}^d (x_{[i]} - y_{[i]})^2/\rho_i^2$, where
$x_{[i]}$ and $y_{[i]}$ denote the elements of $x$ and $y$. The function
$r:\RR\to\RR$ in~(\ref{eq:cov-gen-form}) is the stationary correlation
function of $Z$. From now on, the vector of model parameters will be
denoted by
$\theta = (\sigma^2,\,\rho_1,\,\ldots,\, \rho_d,\ldots,
\sigma_{\varepsilon}^2)\tr \in \Theta \subset \RR^p$, and the
corresponding covariance matrix
$\mb{K}(\x_n,\x_n) + \sigma_{\varepsilon}^2 \mb{I}_{n}$ by
$\mb{K}_{\theta}$.

\begin{table}[t]
  \caption{Some kernel functions available in GPy.
    The Matérn kernel is recommended by
    \cite{stein1999interpolation}. $\Gamma$ denotes the gamma function,
    $\mathcal{K}_{\nu}$ is the modified Bessel function of the second
    kind.} 
  \label{table:kernels}
\begin{center}
\begin{small}
\begin{sc}
\begin{tabular}{p{5cm}l}
\toprule
Kernel              & $r(h)$, $h \in [0, +\infty)$\\
\midrule
{\scriptsize Squared exponential} & $\exp(-\frac{1}{2}r^2)$\\
{\scriptsize Rational Quadratic}  & $(1 + r^2)^{-\nu}$\\ 
{\scriptsize Matérn with param. $\nu>0$}  &
                              $\frac{2^{1-\nu}}{\Gamma(\nu)}\bigg(\sqrt{2\nu}
                              r \bigg)^{\nu}
                              \mathcal{K}_{\nu}\bigg(\sqrt{2\nu} r
                              \bigg)$ \\
\bottomrule
\end{tabular}
\end{sc}
\end{small}
\end{center}
\EOFvspace
\end{table}

\mysubsec{Maximum likelihood estimation}
\label{subsec:mle-parameters}

In this article, we focus on GP implementations where the parameters
$(\theta, \mu)\in\Theta\times \RR$ of the process $Z$ are estimated by
maximizing the likelihood $\mathscr{L}(\Z_n|\theta, \mu)$ of
$\Z_n = (Z_1,\ldots, Z_n)\tr$, or equivalently, by minimizing the
negative log-likelihood~(NLL)
\begin{equation}
     \label{eq:likelihood}
 - \log (\mathscr{L}(\Z_n|\theta, \mu)) \,=\; \frac{1}{2}(\Z_n-\mu\one_n)^{\top}
                                        \mb{K}_{\theta}^{-1}(\Z_n-\mu\one_n)
 +\frac{1}{2}\log \abs{\mb{K}_{\theta}} + \text{constant}.
\end{equation}

This optimization is typically performed by gradient-based methods,
although local maxima can be of significant concern as the likelihood is
often non-convex.  Computing the likelihood and its gradient with
respect to $(\theta, \mu)$ has a $O(n^3 + dn^2)$ computational cost
\citep{rasmussen06:_gauss_proces_machin_learn, petit20:_towar_gauss}.

\mysec{Numerical noise}
\label{sec:numerical-aspects}

\begin{figure}[t]
\begin{center}
\includegraphics[height=5cm]{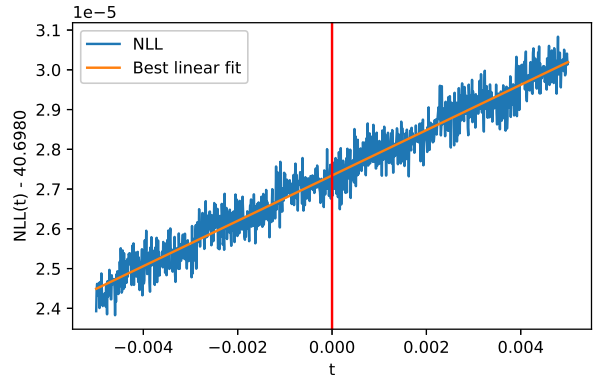}
\caption{Noisy NLL profile along a particular direction in the
  parameter space, with a best linear fit (orange line). This example
  was obtained with GPy while estimating the parameters of a
  Matérn $5/2$ covariance, using $20$ data points sampled from a
  \emph{Branin function}, and setting $\sigma_{\varepsilon}^2 = 0$.  The
  red vertical line indicates the location where the optimization of the
  likelihood stalled.}
\label{fig:noisy-likelihood}
\end{center}
\EOFvspace
\end{figure}

The evaluation of the NLL as well as its gradient
is subject to numerical noise, which can prevent proper convergence of
the optimization algorithms. Figure~\ref{fig:noisy-likelihood} shows a typical
situation where the gradient-based
optimization algorithm stops before converging to an actual minimum. %
In this section, we provide an analysis on the numerical
noise on the NLL using the concept of local condition numbers. We
also show that the popular solution of adding \emph{jitter} cannot be
considered as a fully satisfactory answer to the problem of numerical noise.

Numerical noise stems from both terms of the NLL, namely
$\frac{1}{2}\Z_n^{\top} \mb{K}_{\theta}^{-1}\Z_n$ and
$\frac{1}{2}\log \abs{\mb{K}_{\theta}}$. (For simplification, we assume
$\mu=0$ in this section.)

First, recall that the condition number $\kappa(\mb{K}_\theta)$
of~$\mb{K}_\theta$, defined as the ratio
$\abs{\lambda_{\max}/\lambda_{\min}}$ of the largest eigenvalue to the
smallest eigenvalue \citep{Press_1992}, is the key element for analyzing
the numerical noise on $\mb{K}_{\theta}^{-1}\Z_n$. In double-precision
floating-point approximations of numbers, $\Z_n$ is corrupted by an
error $\eps$ whose magnitude is such that
$\ns{\epsilon}/\ns{\Z_n} \simeq 10^{-16}$.  Worst-case alignment of
$\Z_n$ and $\epsilon$ with the eigenvectors of $\mb{K}_{\theta}$ gives
\begin{equation}\label{eq:condition-number-effect}
\frac{\ns{\mb{K}_{\theta}^{-1} \epsilon}}{\ns{\mb{K}_{\theta}^{-1} \Z_n}}
\simeq \kappa(\mb{K}_{\theta}) \times 10^{-16}\,,
\end{equation}
which shows how the numerical noise is amplified when
$\mb{K}_{\theta}$~becomes ill-conditioned.

The term $\log\abs{\mb{K}_{\theta}}$ is nonlinear in $\mb{K}_{\theta}$,
but observe, using the identity
$\ddiff \log\abs{\mb{K}_{\theta}}/ \ddiff \mb{K}_{\theta} =
\mb{K}_{\theta}^{-1}$, that the differential of $\log \abs{\,\cdot\,}$
at $\mb{K}_{\theta}$ is given by
$H \mapsto \mathrm{Trace}(\mb{K}_{\theta}^{-1} H)$. Thus, the induced
operator norm with respect to the Frobenius norm $\ns{\,\cdot\,}_F$ is
$\ns{\mb{K}_{\theta}^{-1}}_F$. We can then apply results from
\cite{trefethen97} to get a local condition number of the mapping
$\mb{A} \mapsto \log\abs{\mb{A}}$ at~$\mb{K}_{\theta}$:
\begin{equation}
  \kappa( \log \abs{\,\cdot\,},\,  \mb{K}_{\theta})
                            \triangleq \lim_{\epsilon \to 0}
                             \sup_{\ns{\delta_{\mb{A}}}_F \leq \epsilon}
                             \frac{\bigl| \log\abs{\mb{K}_{\theta}
                             +
                             \delta_{\mb{A}}}
                             -
                             \log\abs{\mb{K}_{\theta}} \bigr|
                             }{\bigl|\log\abs{\mb{K}_{\theta}}\bigr|} \frac{\ns{\mb{K}_{\theta}}_F}{\ns{\delta_{\mb{A}}}_F}                       
                             = \frac{\sqrt{\sum_{i=1}^n \frac{1}{\lambda_i^2}} \sqrt{\sum_{i=1}^n \lambda_i^2}}
                             {\abs{\sum_{i=1}^n \log(\lambda_i)}}
  \label{eq:log-det-condition-number}
\end{equation}
where $\lambda_1, \cdots, \lambda_n$ are the (positive) eigenvalues of
$\mb{K}_{\theta}$.  Then, we have
\begin{equation}\label{eq:condition-numbers-link}
\frac{\kappa(\mb{K}_{\theta})}{\abs{ \sum_{i=1}^n \log(\lambda_i)} } \leq \kappa(\log \abs{\,\cdot\,},\, \mb{K}_{\theta}) \leq \frac{n \kappa(\mb{K}_{\theta})}{\abs{\sum_{i=1}^n \log(\lambda_i)} },
\end{equation}
which shows that numerical noise on $\log\abs{\mb{K}_{\theta}}$ is
linked to the condition number of $\mb{K}_{\theta}$.

The local condition number of the quadratic form
$\frac{1}{2}\Z_n\tr\mb{K}_{\theta}^{-1}\Z_n$ as a function of $\Z_n$ can
also be computed analytically.  Some straightforward calculations show
that it is bounded by $\kappa(\mb{K}_{\theta})$.

(When the optimization algorithm stops in the example of
Figure~\ref{fig:noisy-likelihood}, we have
$\kappa(\mb{K}_{\theta}) \simeq 10^{11}$ and
$\kappa(\log \abs{\,\cdot\,},\, \mb{K}_{\theta}) \simeq 10^{9.5}$.  The
empirical numerical fluctuations are measured as the residuals of a
local second-order polynomial best fit, giving noise levels $10^{-7}$,
$10^{-8}$ and $10^{-7.5}$ for $\mb{K}_{\theta}^{-1}\Z_n$,
$\frac{1}{2}\Z_n\tr\mb{K}_{\theta}^{-1}\Z_n$ and
$\log \abs{\mb{K}_{\theta}}$ respectively.  These values are consistent
with the above first-order analysis.)

Thus, when $\kappa(\mb{K}_{\theta})$ becomes large in the course of the
optimization procedure, numerical noise on the likelihood and its
gradient may trigger an early stopping of the optimization algorithm
(supposedly when the algorithm is unable to find a proper direction of
improvement).  It is well-known that $\kappa(\mb{K}_{\theta})$ becomes
large when $\sigma_{\varepsilon}^2 = 0$ and one of the following
conditions occurs: 1) data points are close, 2) the covariance is very
smooth (as for instance when considering the squared exponential
covariance), 3) when the range parameters $\rho_i$ are large. These
conditions arise more often than not. Therefore, the problem of
numerical noise in the evaluation of the likelihood and its gradient is
a problem that should not be neglected in GP implementations.

The most classical approach to deal with ill-conditioned covariance
matrices is to add a small positive number on the diagonal of the
covariance matrix, called \emph{jitter}, which is equivalent to assuming
a small observation noise with variance $\sigma_{\varepsilon}^2>0$.  In
GPy for instance, the strategy consists in always setting a minimal
jitter of $10^{-8}$, which is automatically increased by an amount
ranging from $10^{-6} \sigma^2$ to $10^{-1} \sigma^2$ whenever the
Cholesky factorization of the covariance matrix fails (due to numerical
non-positiveness). The smallest jitter making $\mb{K}_{\theta}$
numerically invertible is kept and an error is thrown if no jitter
allows for successful factorization. However, note that large values for
the jitter may yield smooth, non-interpolating approximations, with
possible unintuitive and undesirable effects
\citep[see][]{andrianakis12:_gauss}, and causing possible convergence
problems in Bayesian optimization.

Table~\ref{table:jitter-influence} illustrates the behaviour of GP interpolation when
$\sigma_{\varepsilon}^2$ is increased.  It appears that finding a
satisfying trade-off between good interpolation properties and low
numerical noise level can be
difficult. Table~\ref{table:jitter-influence} also supports the
connection in \eqref{eq:condition-number-effect} and
\eqref{eq:condition-numbers-link} between noise levels and
$\kappa(\mb{K}_{\theta})$. In view of the results of Figure~\ref{fig:LOO-plots} based on the
default settings of GPy and Table~\ref{table:jitter-influence}, we
believe that adaptive jitter cannot be considered as a do-it-all
solution.

\begin{table*}
  \caption{Influence of the jitter on the GP model (same setting as in
    Figure~\ref{fig:noisy-likelihood}). The table reports the
    condition numbers $\kappa(\mb{K}_{\theta})$ and
    $\kappa(\log \abs{\,\cdot\,},\, \mb{K}_{\theta})$, and the impact on
    the relative empirical standard deviations $\delta_{\rm quad}$ and
    $\delta_{\rm logdet}$ of the numerical noise on $\Z_n^T \mb{K}_{\theta}^{-1} \Z_n$ and
    $\log\abs{\mb{K}_{\theta}}$ respectively (measured using
    second-order polynomial regressions). As $\sigma_{\varepsilon}$
    increases, $\delta_{\rm quad}$ and $\delta_{\rm logdet}$ decrease
    but the interpolation error
    $\sqrt{\mathrm{SSR}/\mathrm{SST}} = \sqrt{\frac{1}{n} \sum_{j = 1}^n
      (Z_j - \hat{Z}_n(x_j))^2} / \mathrm{std}(Z_1, ..., Z_n)$ and
    the NLL increase.  Reducing numerical noise while keeping good interpolation
    properties requires careful attention in practice. }
  \label{table:jitter-influence}
  \vskip -0.4cm
\begin{center}
\footnotesize
\begin{tabular}{| c | >{\centering\arraybackslash}m{1.6cm} | >{\centering\arraybackslash}m{1.6cm} | >{\centering\arraybackslash}m{1.6cm} | >{\centering\arraybackslash}m{1.6cm} | >{\centering\arraybackslash}m{1.6cm} |}
$\sigma_{\varepsilon}^2 ~/~ \sigma^2$ & $0.0$ & $10^{-8}$ & $10^{-6}$ & $10^{-4}$ & $10^{-2}$ \\
  \hline
  \rule{0pt}{10pt}
  $\kappa(\mb{K}_{\theta})$ & $10^{11}$ & $10^{9}$ & $10^{7.5}$ & $10^{5.5}$ & $10^{3.5}$ \\[6pt]

  $\kappa(\log \abs{\,\cdot\,},\, \mb{K}_{\theta})$ & $10^{9.5}$ & $10^{8.5}$ & $10^{6.5}$ & $10^{4.5}$ & $10^{2.5}$ \\[4pt]
  
  $\delta_{\rm quad}$   & $10^{-8}$ \scriptsize $(=10^{11-19})$
                      & $10^{-9.5}$ \scriptsize $(=10^{9-18.5})$
                      & $10^{-10.5}$ \scriptsize $(=10^{7.5-18})$
                      & $10^{-12}$ \scriptsize $(=10^{5.5-17.5})$
                      & $10^{-14}$ \scriptsize $(=10^{3.5-17.5})$ \\[2pt]
  
  $\delta_{\rm logdet}$  & $10^{-7.5}$ \scriptsize $(=10^{9.5-17})$
                      & $10^{-9}$ \scriptsize $(=10^{8.5-17.5})$
                      & $10^{-11}$ \scriptsize $(=10^{6.5-17.5})$
                      & $10^{-13.5}$ \scriptsize $(=10^{4.5-18})$
                      & $10^{-15.5}$ \scriptsize $(=10^{2.5-18})$\\[2pt]

  $- \log (\mathscr{L}(\Z_n|\theta))$ & $40.69$ & $45.13$ & $62.32$ & $88.81$ & $124.76$ \\[6pt]

  $\sqrt{\mathrm{SSR}/\mathrm{SST}}$  & $3.3 \cdot 10^{-10}$ & $1.2 \cdot 10^{-3}$ & $0.028$ & $0.29$ & $0.75$
\end{tabular}
\end{center}
\EOFvspace
\end{table*}

\mysec{Strategies for improving likelihood maximization}
\label{sec:healing}

In this section we investigate simple but hopefully efficient levers /
strategies to improve available implementations of MLE for GP interpolation, beyond
the control of the numerical noise on the likelihood using jitter. %
We mainly focus on %
1) initialization methods for the optimization procedure, %
2) stopping criteria, %
3) the effect of ``restart'' strategies and %
4) the effect of the parameterization of the covariance.

\mysubsec{Initialization strategies}
\label{sec:init-methods}

Most GP implementations use a gradient-based local optimization
algorithm to maximize the likelihood that requires the specification
of starting/initial values for the parameters. %
In the following, we consider different initialization strategies.

\mypar{Moment-based initialization.} %
A first strategy consists in setting the parameters using empirical
moments of the data. %
More precisely, assuming a constant mean $m = \mu$, and a stationary
covariance $k$ with variance~$\sigma^2$ and range parameters
$\rho_1,\,\ldots,\,\rho_d$, set
\begin{eqnarray}
  \mu_{\mathrm{init}} &=& \mathop{\mathrm{mean}}\, (Z_1,\, \hdots,\,Z_n), \\
  \sigma^2_{\mathrm{init}} &=& \mathop{\mathrm{var}}\, (Z_1,\, \hdots,\,Z_n),
  \label{eqref:classic-var-init}\\
  \rho_{k,\, \mathrm{init}}  &=& \mathop{\mathrm{std}}\,
  (x_{1,\,[k]},\,\hdots,\,x_{n,\,[k]}),\quad k=1,\, \ldots,\,d,~
                          \label{eqref:length-scale-init}
\end{eqnarray}
where $\mathrm{mean}$, $\mathrm{var}$ and $\mathrm{std}$ stand for the
empirical mean, variance and standard deviation, and $x_{i,\,[k]}$
denotes the $k$th coordinate of~$x_i\in\RR^d$. %
The rationale behind~\eqref{eqref:length-scale-init} \citep[following,
e.g.,][]{rasmussen06:_gauss_proces_machin_learn} is that the range
parameters can be thought of as the distance one has to move in the
input space for the function value to change significantly and we
assume, a priori, that this distance is linked to the dispersion of data
points.
  
In GPy for instance, the default initialization consists in setting
$\mu = 0$, $\sigma^2 = 1$ and $\rho_k = 1$ for all $k$. %
This is equivalent to the \emph{moment-based} initialization scheme when
the data (both inputs and outputs) are centered and standardized. %
The practice of standardizing the input domain into a unit length
hypercube has been proposed \citep[see, e.g.,][]{snoek2012} to deal with
numerical issues that arise due to large length scale values.

\mypar{Profiled initialization.} %
Assume the range parameters $\rho_1,\,\ldots,\, \rho_d$ (and more
generally, all parameters different from $\sigma^2$,
$\sigma_{\varepsilon}^2$ and $\mu$) are fixed, and set
$\sigma_{\varepsilon}^2 = \alpha \sigma^2$, with a prescribed
multiplicative factor $\alpha\geq 0$. %
In this case, the NLL can be optimized analytically w.r.t. $\mu$ and
$\sigma^2$. %
Optimal values turn out to be the generalized least squares solutions
\begin{align}
  \mu_{\mathrm{GLS}}
  &= (\one_n\tr \mb{K}_{\tilde{\theta}}^{-1} \one_n)^{-1}\one_n\tr \mb{K}_{\tilde{\theta}}^{-1}\Z_n\,,
      \label{eq:analytical-mean}\\
  \sigma_{\mathrm{GLS}}^2
  &= \frac{1}{n} (\Z_n - \mu_{\mathrm{GLS}}\mskip 3mu \one_n)\tr
      \mb{K}_{\tilde{\theta}}^{-1} (\Z_n - \mu_{\mathrm{GLS}}\mskip 3mu \one_n)\,,
      \label{eq:analytical-var}
\end{align}
where
$\tilde{\theta} = (\sigma^2,\,\rho_1,\,\ldots,\, \rho_d,\ldots,\,
\sigma_{\varepsilon}^2)\tr\in\Theta$, with $\sigma^2=1$ and
$\sigma_{\varepsilon}^2= \alpha$. %
Under the \emph{profiled} initialization scheme,
$\rho_{1}, \ldots, \rho_{d}$ are set
using~\eqref{eqref:length-scale-init}, $\alpha$ is prescribed
according to user's preference, and $\mu$ and $\sigma^2$ are initialized
using \eqref{eq:analytical-mean} and \eqref{eq:analytical-var}.

\mypar{Grid-search initialization.} %
\emph{Grid-search} initialization is a \emph{profiled} initialization
with the addition of a grid-search optimization for the range
parameters.

Define a nominal range vector~$\rho_0$ such that
\begin{equation*}
  \rho_{0,[k]} \;=\; \sqrt{d}\, \left(
    \max_{1 \leq i \leq n} x_{i,[k]} - \min_{1 \leq i \leq n} x_{i,[k]}
  \right),\quad 1 \le k \le d.
\end{equation*}
Then, define a one-dimensional grid of size $L$ (e.g., $L=5$) by
taking range vectors proportional to $\rho_0$: %
$\{\alpha_1 \rho_0,\, \ldots,\, \alpha_L \rho_0\}$, where the
$\alpha_i$s range, in logarithmic scale, from a ``small'' value (e.g.,
$\alpha_1 = 1/50$) to a ``large'' value (e.g., $\alpha_L = 2$). %
For each point of the grid, the likelihood is optimized with respect to
$\mu$ and $\sigma^2$ using~\eqref{eq:analytical-mean}
and~\eqref{eq:analytical-var}. %
The range vector with the best likelihood value is selected. %
(Note that this initialization procedure is the default initialization
procedure in the Matlab/GNU Octave toolbox STK.)

\mysubsec{Stopping condition}

Most GP implementations rely on well-tested gradient-based optimization
algorithms.  For instance, a popular choice in Python implementations is
to use the limited-memory BFGS algorithm with box constraints
\citep[L-BFGS-B; see][]{byrd1995limited} of the SciPy ecosystem. (Other
popular optimization algorithms include the ordinary BFGS, truncated
Newton constrained, SQP, etc.; see, e.g., \cite{NoceWrig06}.) The
L-BFGS-B algorithm, which belongs to the class of quasi-Newton
algorithms, uses limited-memory Hessian approximations and shows good
performance on non-smooth functions \citep{curtis15:_newton}.

Regardless of which optimization algorithm is chosen, the user usually
has the possibility to tune the behavior of the optimizer, and in
particular to set the stopping condition. Generally, the stopping
condition is met when a maximum number of iterations is reached or when
a norm on the steps and/or the gradient become smaller than a threshold.

By increasing the strictness of the stopping condition during the
optimization of the likelihood, one would expect better parameter
estimations, provided the numerical noise on the likelihood does not
interfere too much.

\mysubsec{Restart and multi-start strategies}
\label{subsec:restart}

Due to numerical noise and possible non-convexity of the likelihood with
respect to the parameters, gradient-based optimization algorithms may
stall far from the global optimum. %
A common approach to circumvent the issue is to carry out several
optimization runs with different initialization points. %
Two simple strategies can be compared.

\mypar{Restart.} %
In view of Figure~\ref{fig:noisy-likelihood}, a first simple strategy is
to restart the optimization algorithm to clear its memory (Hessian
approximation, step sizes\ldots), hopefully allowing it to escape a
possibly problematic location using the last best parameters as initial
values for the next optimization run. %
The optimization can be restarted a number of times, until a budget
$N_{\mathrm{opt}}$ of restarts is spent or the best value for the
likelihood does not improve.

\mypar{Multi-start.} %
Given an initialization point
$(\theta_{\mathrm{init}}, \mu_{\mathrm{init}})\in\Theta\times\RR$, a
multi-start strategy consists in running $N_{\mathrm{opt}} > 1$
optimizations with different initialization points corresponding to
perturbations of the initial point
$(\theta_{\mathrm{init}}, \mu_{\mathrm{init}})$. %
In practice, we suggest the following rule for building the
perturbations: first, move the range parameters around
$(\rho_{1,\,\mathrm{init}},\, \ldots,\, \rho_{d,\,\mathrm{init}})^T$
(refer to Section~\ref{sec:numerical-experiments} for an
implementation); then, propagate the perturbations on $\mu$ and
$\sigma^2$ using~(\ref{eq:analytical-mean})
and~(\ref{eq:analytical-var}). %
The parameter with the best likelihood value over all optimization
runs is selected.

\mysubsec{Parameterization of the covariance function}

The parameters of the covariance functions are generally positive real
numbers ($\sigma^2$, $\rho_1, \rho_2\ldots$) and are related to scaling
effects that act ``multiplicatively'' on the predictive
distributions. Most GP implementations introduce a reparameterization 
using a monotonic one-to-one mapping $\tau:\RR_+^{\star} \to \RR$, 
acting component-wise on the positive parameters of $\theta$, resulting 
in a mapping $\tau:\Theta\to \Theta^{\prime}$. Thus, for carrying out 
MLE, the actual criterion $J$ that is optimized in most implementations 
may then be written as
\begin{equation}
  \label{eq:NLL-transf}
  J: \theta^{\prime}\in\Theta^{\prime} \mapsto - \log (\mathscr{L}(\Z_n|\tau^{-1}(\theta^{\prime}), c))\,.
\end{equation}
Table~\ref{table:parametrizations} lists two popular
reparameterization mappings $\tau$.

\begin{figure}
\centering
\includegraphics[height=5cm]{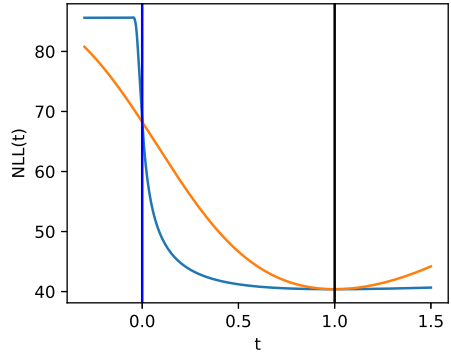}
\vspace*{-1.4em}
\caption{Profiles of the NLL along a linear path $t$ through the
  \emph{profiled} initialization point (at zero, blue vertical line) and
  the optimum (at one, black vertical line). Orange (resp. blue) line
  corresponds to the \emph{log} (resp. \emph{invsoftplus})
  reparameterization.}
\label{fig:parametrization}
\EOFvspace
\end{figure}

The effect of reparameterization is to ``reshape'' the likelihood.
Typical likelihood profiles using the \emph{log} and the so-called \emph{invsoftplus}
reparameterizations are shown on
Figure~\ref{fig:parametrization}. Notice that the NLL may be almost flat
in some regions depending on the reparameterization. Changing the shape
of the optimization criterion, combined with numerical noise, may or may
not facilitate the convergence of the optimization.

\begin{table}[t]
  \caption{Two popular reparameterization mappings $\tau$, as
    implemented, for example, in GPy and STK
    respectively. For \emph{invsoftplus}, notice
    parameter $s>0$, which is introduced when input standardization is
    considered (see Section~\ref{sec:numerical-experiments}).}
  \label{table:parametrizations}
\begin{center}
\begin{small}
\begin{sc}
\begin{tabular}{p{3cm} >{\centering\arraybackslash}m{3cm} c}
\toprule
Reparam. method & $\tau: \RR_{+}^{\star}\to \RR$ & $\tau^{-1}: \RR \to \RR_{+}^{\star}$\\
\midrule
invsoftplus$(s)$ & $\log(\exp(\theta/s)-1)$ & $s\log(\exp(\theta')+1)$\\ 
log & $\log(\theta)$ & $\exp(\theta')$\\
\bottomrule
\end{tabular}
\end{sc}
\end{small}
\end{center}
\EOFvspace
\end{table}

\mysec{Numerical study}
\label{sec:numerical-experiments}

\mysubsec{Methodology}
\label{subsec:ECDF-construction}

The main metric used in this numerical study is based on empirical
cumulative distributions (ECDFs) of differences on NLL values.

More precisely, consider $N + 1$ optimization schemes
$S_0, S_1,\ldots, S_N$, where $S_0$ stands for a ``brute-force''
optimization scheme based on a very large number of multi-starts,
which is assumed to provide a robust MLE, %
and $S_1,\ldots, S_N$ are optimization schemes to be compared. %
Each optimization scheme is run on $M$ data sets~$D_j$,
$1 \le j \le M$, and we denote by $e_{i,\,j}$ the difference
\begin{equation*}
  e_{i,j} = \mathrm{NLL}_{i,\,j} - \mathrm{NLL}_{0,\, j}\,,
  \quad 1 \le i \le N, \quad 1 \le j \le M,
\end{equation*}
where $\mathrm{NLL}_{i,j}$ the NLL value obtained by optimization
scheme $S_i$ on data set $D_j$.

A good scheme $S_i$ should concentrate the empirical distribution of
the sample $E_i = \{e_{i,j}, j=1,\ldots,\, M\}$ around zero---in other
words, the ECDF is close to the ideal CDF
$e \mapsto \one_{[0,\infty[}(e)$. %
Using ECDF also provides a convenient way to compare performances: a
strategy with a ``steeper'' ECDF, or larger area under the ECDF, is
better.

\mysubsec{Optimization schemes}
\label{sec:opt-schemes}

All experiments are performed using GPy version $1.9.9$, with the
default L-BFGS-B algorithm. %
We use a common setup and vary the configurations of the optimization
levers as detailed below.

\mypar{Common setup.} %
All experiments use an estimated constant mean-function, an anisotropic
Matérn covariance function with regularity~$\nu = 5/2$, and
we assume no observation noise (the adaptive jitter of GPy ranging from
$10^{-6} \sigma^2$ to $10^{2} \sigma^2$ is used, however).

\mypar{Initialization schemes.} %
Three initialization procedures from
Section~\ref{sec:init-methods} are considered.

\mypar{Stopping criteria.} %
We consider two settings for the stopping condition of the L-BFGS-B
algorithm, called \emph{soft} %
(the default setting: \texttt{maxiter}$=1000$,
\texttt{factr}=$10^{7}$, \texttt{pgtol}$10^{-5}$) %
and \emph{strict} %
(\texttt{maxiter}$=1000$, \texttt{factr}=$10$,
\texttt{pgtol}$=10^{-20}$).

\mypar{Restart and multi-start.} %
The two strategies of Section~\ref{subsec:restart} are implemented
using a \emph{log} reparameterization and initialization points
$(\theta_{\mathrm{init}}, \mu_{\mathrm{init}})$ determined using a
\emph{grid-search} strategy. %
For the \emph{multi-start} strategy the initial range parameters are
perturbed according to the rule
$\rho \leftarrow \rho_{\mathrm{init}} \cdot 10^{\eta}$ where $\eta$ is drawn
from a $\Ncal(0, \sigma_{\eta}^2)$ distribution. We take $\sigma_{\eta} = 
\log_{10}(5) / 1.96~ (\approx 0.35)$, to ensure that about 0.95 of the
distribution of $\rho$ is in the interval  $\left[  1/5 \cdot
  \rho_{\mathrm{init}},~5\cdot \rho_{\mathrm{init}}\right]$.

\mypar{Reparameterization.} %
We study the \emph{log} reparameterization and two variants of the
\emph{invsoftplus}. %
The first version called \emph{no-input-standardization} simply
corresponds to taking $s=1$ for each range parameter. %
The second version called \emph{input-standardization} consists in
scaling the inputs to a unit standard deviation on each dimension (by
taking the corresponding value for~$s$).

\mysubsec{Data sets}
\label{subsec:datasets}

The data sets are generated from six well-known test functions in the
literature of Bayesian optimization: %
the Branin function \citep[$d=2$; see,
e.g.][]{surjanovic13:_virtual_librar_simul_exper}, %
the Borehole function \citep[$d=8$; see, e.g.][]{worley87}, %
the Welded Beam Design function \citep[$d=4$;
see][]{chafekar03:_const_multi_optim_using_stead}, %
the g10 function \citep[$d=8$; see][p. 128]{Hedar04studieson}, %
along with two modified versions, g10mod and g10modmod
\citep[see][]{feliot2017phd}.

Each function is evaluated on Latin hypercube samples with a
multi-dimensional uniformity criterion \citep[LHS-MDU;
][]{deutsch12:_latin}, with varying sample size
$n\in\{3d,\, 5d,\, 10d,\,20d\}$, resulting in a total of
$6\times 4 = 24$ data sets.

\mysubsec{Results and findings}
\label{subsec:results}

\begin{figure}[hb]
  \begin{center}
    \subfigure[effect of reparameterization]{\includegraphics[width=6.8cm]{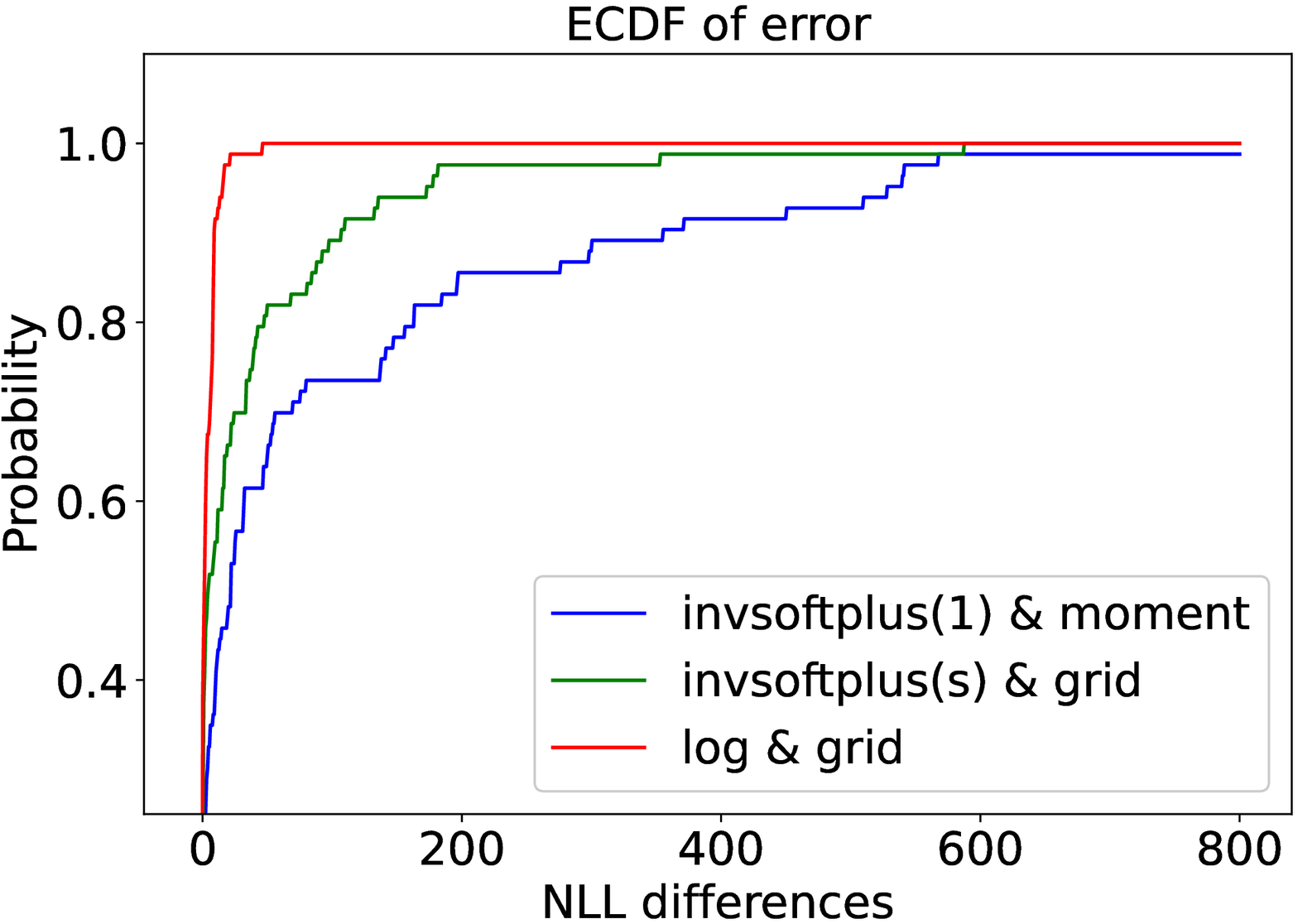}\hspace{-0.5cm}}%
    \subfigure[effect of initialization ]{\includegraphics[width=6.8cm]{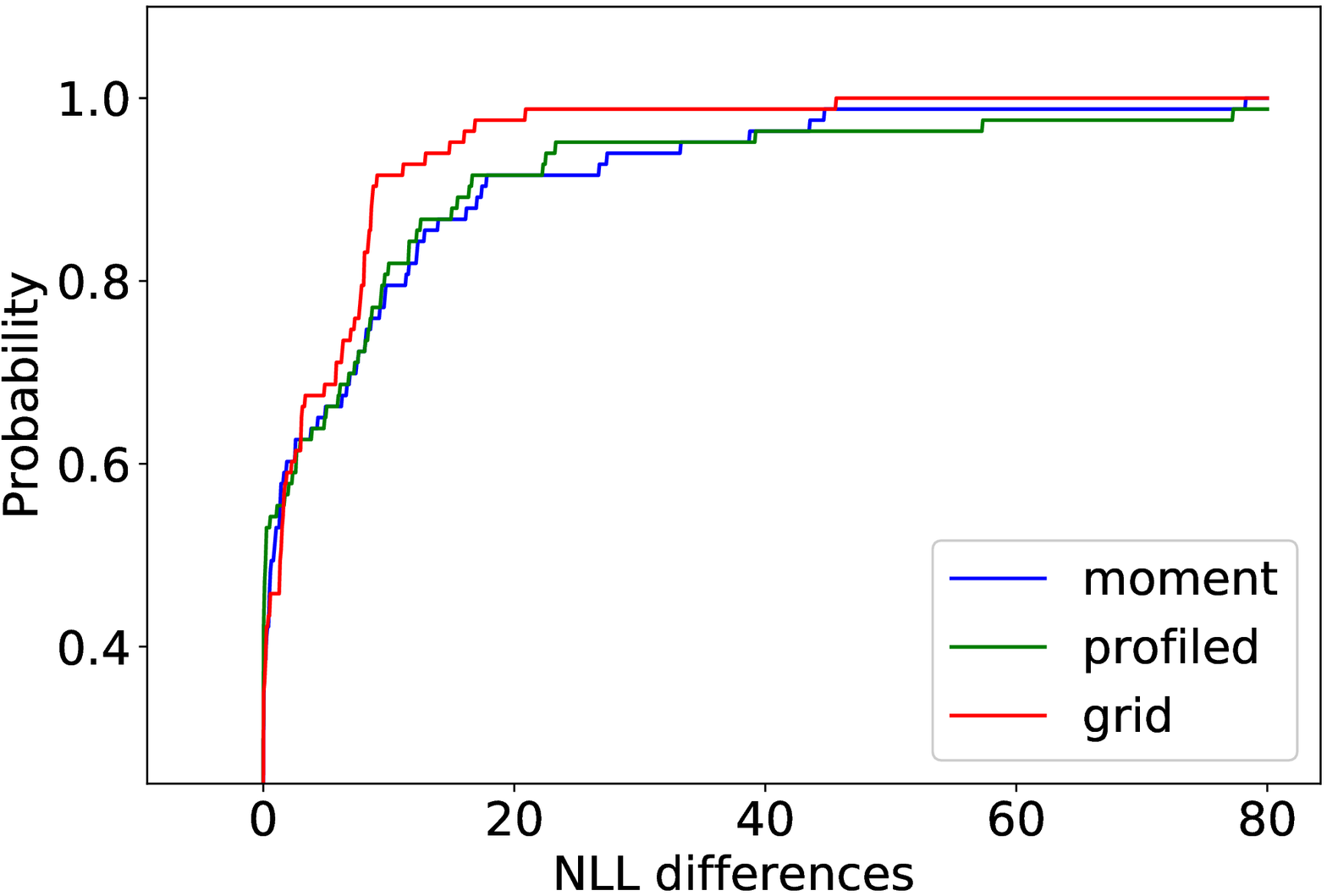}}
    \caption{Initialization and reparameterization methods. (a) ECDFs
      corresponding to the best initialization method for each of the
      three reparameterizations---red line: \emph{log} reparam. with
      \emph{grid-search} init.; green line: \emph{invsoftplus} with
      \emph{input-standardization} reparam. and \emph{grid-search}
      init; blue line: \emph{invsoftplus} with
      \emph{no-input-standardization} reparam. and \emph{moment-based}
      init. (b) ECDFs for different initialization methods for the
      $\log$ reparameterization.}
    \label{fig:paramplots}
  \end{center}
  \EOFvspace
\end{figure}

Figure~\ref{fig:paramplots} shows the effect of reparameterization and
the initialization method. %
Observe that the \emph{log} reparameterization performs significantly
better than the \emph{invsoftplus} reparameterizations. %
For the \emph{log} reparameterization, observe that the
\emph{grid-search} strategy brings a moderate but not negligible gain
with respect to the two other initialization strategies, which behave
similarly.

Next, we study the effect of the different restart strategies and the
stopping conditions, on the case of the \emph{log} reparameterization
and \emph{grid-search} initialization. %
The metric used for the comparison is the area under the ECDFs of the
differences of NLLs, computed by integrating the ECDF between~$0$
and~$\mathrm{NLL}_{\text{max}} = 100$. Thus, a perfect optimization
strategy would achieve an area under the ECDF equal to 100. %
Since the \emph{multi-start} strategy is stochastic, results are
averaged over 50 repetitions of the optimization procedures (for each
$N_{\mathrm{opt}}$ value, the optimization strategy is repeated 50
times). %
The areas are plotted against the computational run time. Run times
are averaged over the repetitions in the case of the
\emph{multi-start} strategy.

\begin{figure}[hb]
  \begin{center}
    \subfigure[restart with $N_{\mathrm{opt}}=1,\,\ldots,\,20$]{\includegraphics[trim=10 0 50 15,clip,width=6.0cm]{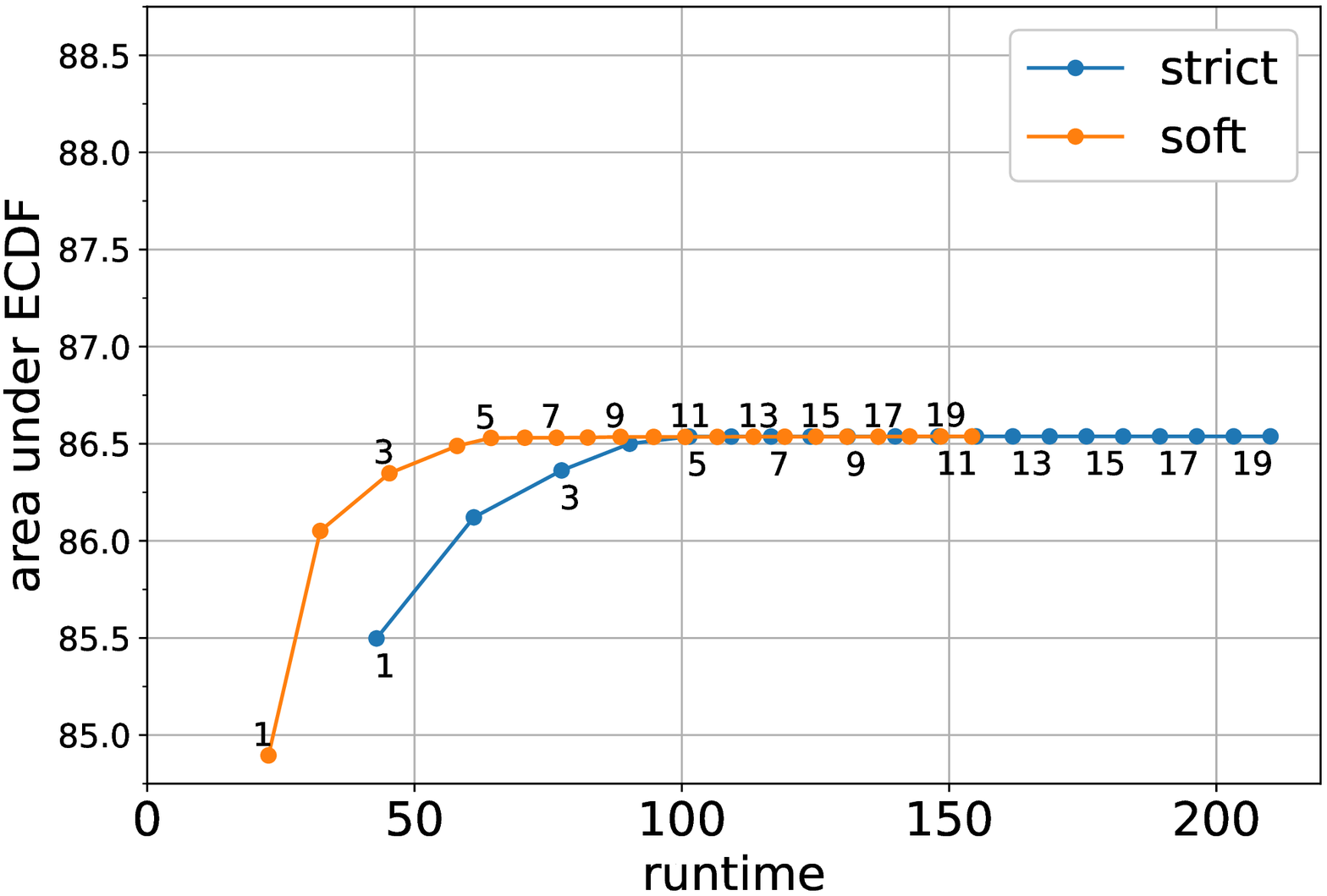}\hspace{-0.cm}}
    \subfigure[multi-start with $N_{\mathrm{opt}}=1,\,\ldots,\,20$, $\sigma_{\eta} = 0.35$]{\includegraphics[trim=10 0 50 15,clip,width=6.0cm]{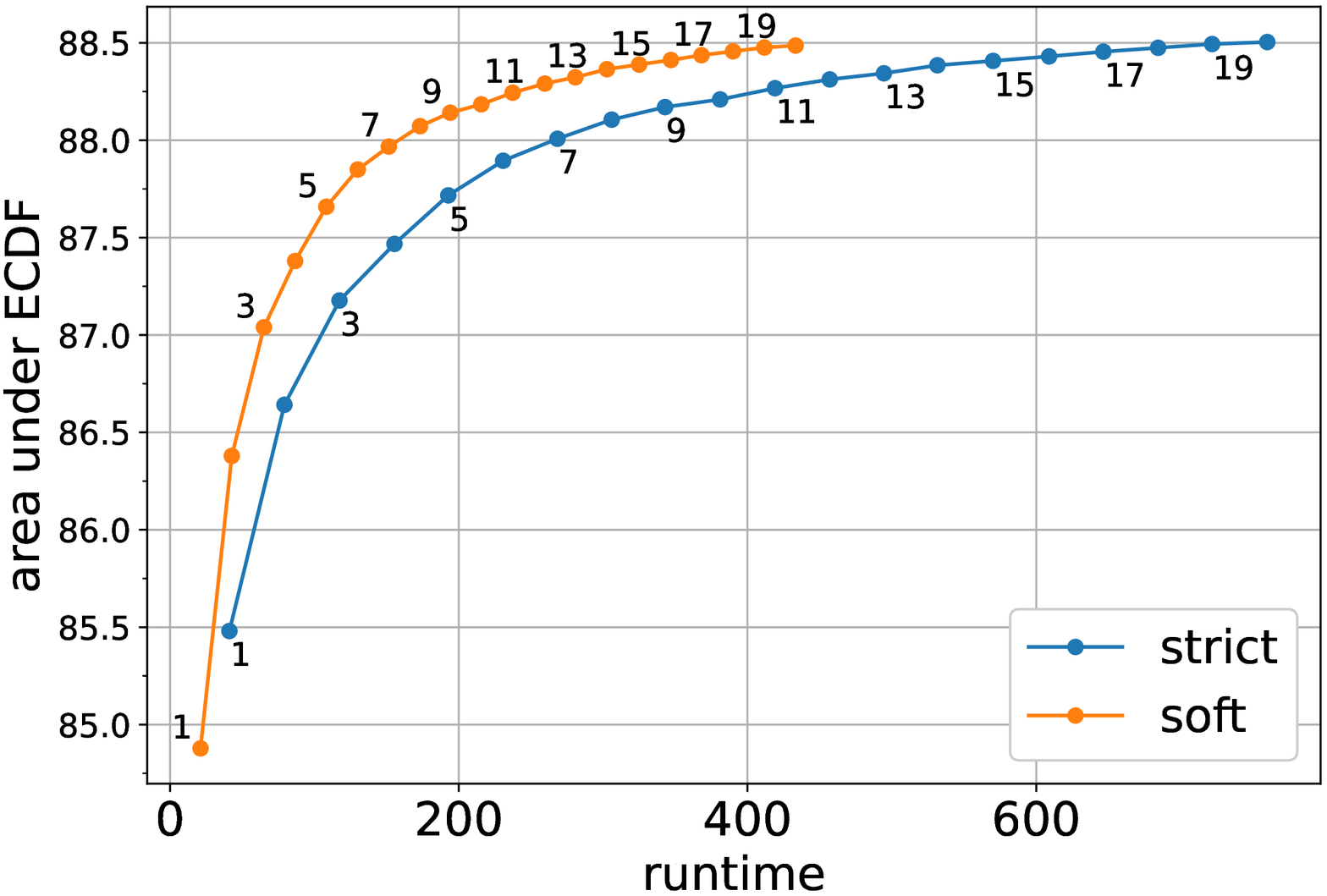}}
    \caption{Area under the ECDF against run time: (a) \emph{restart}
      strategy; (b) \emph{multi-start} strategy. The maximum areas obtained 
	  are respectively $86.538$ and $88.504$.}
    \label{fig:restart}
  \end{center}
  \EOFvspace
\end{figure}

Figure~\ref{fig:restart} shows that the \emph{soft} stopping condition
seems uniformly better. The \emph{restart} strategy yields small
improvements using moderate computational overhead. The \emph{multi-start} strategy is able to
achieve the best results at the price of higher computational costs.

\afterpage{\clearpage}

\vspace{-0.5em}

\mysec{Conclusions and recommendations}
\label{sec:conclusions}

Our numerical study has shown that the parameterization of the
covariance function has the most significant impact on the accuracy of
MLE in GPy. Using \emph{restart} / \emph{multi-start} strategies is also
very beneficial to mitigate the effect of the numerical noise on the
likelihood. The two other levers have second-order but nonetheless
measurable influence.

These observations make it possible to devise a recommended combination
of improvement levers---for GPy at least, but hopefully transferable to
other software packages as well.  When computation time matters, an
improved optimization procedure for MLE consists in choosing the
combination of a \emph{log} reparameterization, with a
\emph{grid-search} initialization, the \emph{soft} (GPy's default)
stopping condition, and a small number, say $N_{\mathrm{opt}}=5$, of
restarts.

Figure~\ref{fig:LOO-plots} and Table~\ref{table:LOO-mse} are based on
the above optimization procedure, which results in significantly
better likelihood values and smaller prediction errors. The
\emph{multi-start} strategy can be used when accurate results are
sought.

Several topics could be investigated in the future: the optimization of alternative selection
criteria, in particular criteria based on leave-one-out procedures,
the case of regression, the problem of parameterization
in relation to that of the identifiability of the parameters \cite[see,
e.g.,][]{anderes10:_gauss}.

As a conclusion, our recommendations are not intended to be universal, but
will hopefully encourage researchers and users to develop and use more
reliable and more robust GP implementations, in Bayesian optimization or
elsewhere.

\vspace{-1em}
%
%
%
%

\bibliographystyle{apalike}
\bibliography{references}
\end{document}